\documentclass[10pt,conference,letterpaper]{IEEEtran}

\usepackage[utf8]{inputenc}
\usepackage{amsmath}
\usepackage{mathtools}
\usepackage{amsthm}
\usepackage{times}
\usepackage{algorithm}
\usepackage[noend]{algorithmic}
\usepackage{graphicx}
\usepackage{verbatim}
\usepackage{amssymb}
\usepackage{cite}
\usepackage{epsfig}
\usepackage{color}
\usepackage{url}
\usepackage{xcolor,colortbl}
\usepackage{multirow}
\usepackage{grffile}
\usepackage{epstopdf}
\usepackage{soul}

\makeatletter
\newcommand*\titleheader[1]{\gdef\@titleheader{#1}}
\AtBeginDocument{%
	\let\st@red@title\@title
	\def\@title{%
		\bgroup\normalfont\small\centering\@titleheader\par\egroup
		\vskip0.35em\st@red@title}
}
\makeatother

\title{Path Planning in Support of Smart Mobility Applications using Generative Adversarial Networks}

\titleheader{Under review as a conference paper at IEEE SmartData-2018}

\begin{document}

\author{\IEEEauthorblockN{Mehdi Mohammadi}
\IEEEauthorblockA{Department of Computer Science\\
Western Michigan University\\
Email: mehdi.mohammadi@wmich.edu}
\and
\IEEEauthorblockN{Ala Al-Fuqaha}
\IEEEauthorblockA{Department of Computer Science\\
Western Michigan University\\
Email: ala.al-fuqaha@wmich.edu}
\and
\IEEEauthorblockN{Jun-Seok Oh}
\IEEEauthorblockA{Department of Civil and Construction\\
Western Michigan University\\
Email: jun.oh@wmich.edu}
}

\IEEEcompsoctitleabstractindextext{
\begin{abstract}
This paper describes and evaluates the use of Generative Adversarial Networks (GANs) for path planning in support of smart mobility applications such as indoor and outdoor navigation applications, individualized wayfinding for people with disabilities (e.g., vision impairments, physical disabilities, etc.), path planning for evacuations, robotic navigations, and path planning for autonomous vehicles. We propose an architecture based on GANs to recommend accurate and reliable paths for navigation applications. The proposed system can use crowd-sourced data to learn the trajectories and infer new ones. The system provides users with generated paths that help them navigate from their local environment to reach a desired location. As a use case, we experimented with the proposed method in support of a wayfinding application in an indoor environment. Our experiments assert that the generated paths are correct and reliable. The accuracy of the classification task for the generated paths is up to 99\% and the quality of the generated paths has a mean opinion score of 89\%.   

\end{abstract}

\begin{IEEEkeywords}
	Path planning, smart mobility, wayfinding, intelligent transportation systems, generative adversarial network, GAN, Internet of Things, smart cities.
\end{IEEEkeywords}
}

\maketitle

\thispagestyle{plain} 
\pagestyle{plain}

\IEEEdisplaynotcompsoctitleabstractindextext
\IEEEpeerreviewmaketitle

\section{Introduction}\label{sec:Introduction}

The development of smart services on top of the Internet of Things (IoT) infrastructure is a movement that aims to introduce new efficiency and comfort applications to our societies; therefore, contributing to livable communities through the internetworking of smart objects \cite{mohammadi2017enabling}. However, the development of such smart services is not straightforward and entails many challenges. One main challenge is the manual labeling of datasets to train such systems; an approach that is very expensive and time-consuming. As reported in \cite{ledlie2012mole}, collecting fingerprinting data for a large office for indoor localization costs up to \$10,000. One alternative solution is crowd-sourcing data. Crowd-sourcing allows end-users to participate in data gathering and annotation tasks through their mobile devices.

In order to develop smart services, Deep Learning (DL) techniques are widely used in IoT and smart city domains for a variety of applications such as Intelligent Transportation Systems (ITS), smart agriculture, smart homes, etc.~\cite{mohammadi2017deep}. Among DL techniques, Generative Adversarial Networks (GANs)~\cite{goodfellow2014generative} are mostly used for vision-based tasks but have been rarely used in the IoT domain. 

Smart mobility refers to the utilization of information and communications technologies (ICT) augmented with artificial intelligence for the optimization and efficient distribution of traffic flows and transportation services \cite{benevolo2016smart}. Smart Mobility is one of the prominent functionalities of a smart city as it can improve the quality of life of almost all the citizens. Path planning plays a significant role for many smart mobility applications. For example, path planning is a major component for emergency evacuation situations in a building or even a city scale. If evacuees have adequate information about the exit ways, smart mobility applications can prevent potential fatalities due to severe crowd congestions or choosing paths that lead to dangerous areas. Numerous approaches have been proposed for intelligent path planning; however, deploying feasible techniques remains a challenge~\cite{chen2015gofast}.

The motivations behind this work are as follows: 
\begin{itemize}
\item Development of indoor path planning and wayfinding technologies for use by disabled commuters through smartphone applications is recommended to improve their mobility and quality of life \cite{legge2016indoor}. These applications can be augmented with multimodal information access to satisfy the needs of different visual and hearing disabilities.

\item In emergency situations and building evacuation scenarios, individualized path planning can be devised in real-time for different groups of people to satisfy their needs. For example,  people using wheelchairs will receive a recommended path to exit that differs from the one recommended to people in the same area that do not have physical disabilities.     

\item Gathering data to train a machine learning system is a time-consuming and expensive process. Crowd-sourced data is a potential alternative to gather the desired data in shorter time and at a lower cost. Since people tend to choose the shortest paths for their navigations between two points, gathering such crowd-sourced data can help to train a machine learning system based on the historical traces of the navigation behavior of other users. 

\end{itemize}

\begin{figure}
	\begin{center}		
		\includegraphics[width=.45\textwidth]{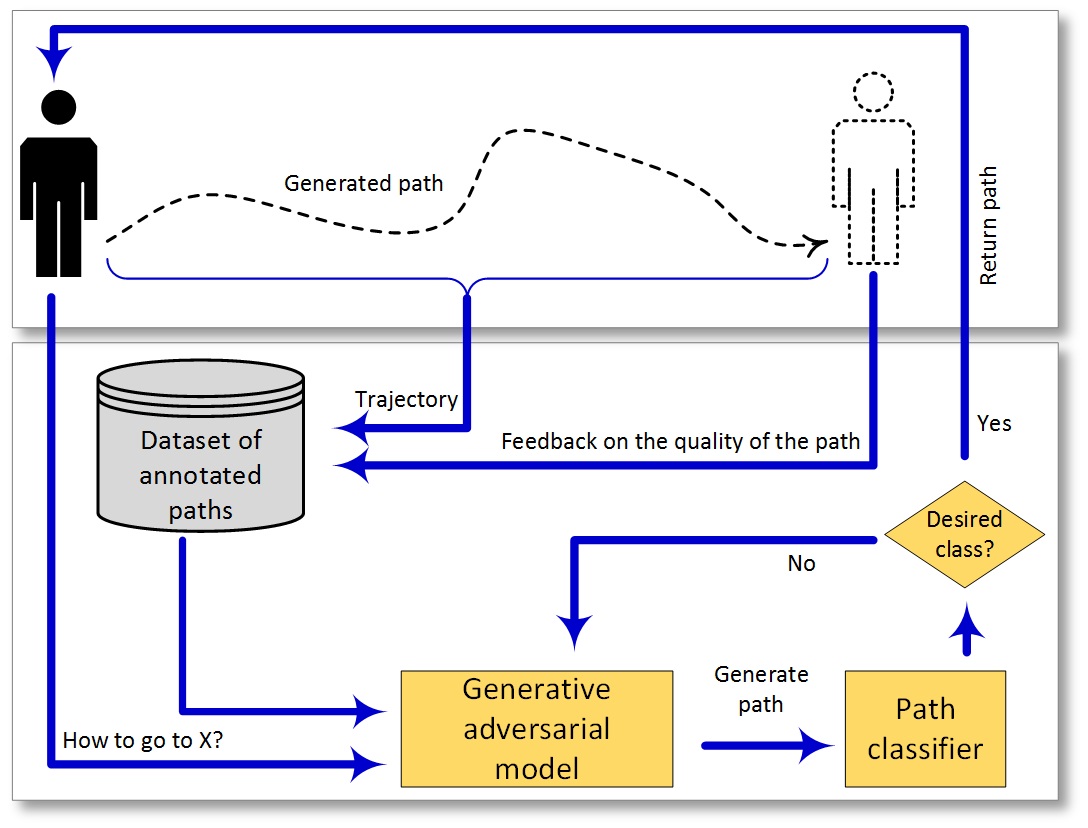}

	\end{center}	
	\caption{The overall architecture of the proposed system.}\label{fig:architecture}	
\end{figure}

The goal of this research is to explore the use of GANs to generate individualized paths that meet the users' needs. To this end, the system accurately determines the location of a user in unfamiliar environments and helps them to navigate to their desired destinations. The system uses a machine learning model that is trained using crowd-sourced data. The system can learn the correct paths through the recorded trajectories of other users, so that in scenarios where no data is recorded between a given source-destination pair, the system can generate a path that is most likely to be a valid and correct path. This way we utilize the experience of other users instead of wasting the collected data. Our method is inspired by the recent successes and advancements in data-driven methods on computer vision problems using generative adversarial networks \cite{radford2015unsupervised,reed2016generative}. Figure~\ref{fig:architecture} illustrates the overall architecture of the system. A GAN model is trained on an existing dataset of annotated trajectories. When a user asks for a path to a specific destination, GAN generates the desired path and reports it to the user. The user then follows the generated path and sends a feedback to the system about the quality of the generated path.

To assess the effectiveness of the proposed approach, we applied it to an indoor path planning application. The indoor environment is equipped with IoT devices deployed in public access areas to help in the localization process of the users. Our results show that the generated paths do not deviate from the safe boundaries (i.e., public access areas) and accurately lead to the desired destinations. 

The main contributions of this paper are:
\begin{enumerate}

\item We propose an approach for path planning that exploits the availability of users' trajectory data and their associated qualitative feedback. At the core of our proposed approach, GAN generates the viable paths that lead to a specified destination. 

\item Our approach does not depend on the availability of pre-engineered training dataset in order for the system to work in real environments. The system can generalize the best solutions by learning from the trajectories of the users over the time. Consequently, the results converge to the optimal ones over time.
  
\end{enumerate}

The rest of the paper is organized as follows. Section \ref{sec:relatedworks} presents several related research studies. Section \ref{sec:method} provides the details of the proposed system followed by section \ref{sec:usecase} that brings the proposed approach to a wayfinding application for the visually impaired. Section \ref{sec:evaluation} presents our experimental results. Finally, section \ref{sec:conclusion} concludes the paper and introduces several directions for potential extensions.   

\section{Related Work}\label{sec:relatedworks}

Many indoor localization and navigation solutions have been proposed based on a wide range of technologies, such as vision, visual light communications (VLC), infrared, sound, WiFi, RFID, and Bluetooth Low Energy (BLE). The cost of the underlying technologies and devices is an important factor to consider when deploying location-based services. Among the aforementioned technologies, BLE is a low-cost technology that has been utilized in recent research works as well as commercial applications. To deploy indoor location-aware systems for environments that are not equipped with wireless networks, the use of the BLE technology (e.g., iBeacon devices) is advantageous. Beyond the low-cost of iBeacons, these devices also can be deployed quickly and easily without changing or even tapping into the building's electrical and communications infrastructure.

One of the most promising approaches to implement location-aware services is based on Relative Signal Strength (RSS) fingerprinting. Such approach needs to address some challenges including fingerprint annotation and device diversity \cite{wang2016indoor}. The use of fingerprint-based approaches to identify an indoor position has been studied well in the past decade. A variety of machine learning approaches have been utilized in this context including Support Vector Machines (SVM), K-Nearest Neighbors (KNN), Bayesian-based filtering, transfer learning, and Deep Neural Networks (DNNs) \cite{zhang2016deep}.

Li \textit{et al.} \cite{li2016self} proposed an indoor navigation system based on off-the-shelf smartphone sensors and magnetic features. They used several approaches to enhance the accuracy of the system including multi-dimensional dynamic time warping, weighted k-nearest neighbors, and exploiting magnetic gradient fingerprints. They also mitigated the impact of magnetic matching mismatches to reduce  the position errors. They reported root mean square error between 4.3 m and 5.6 m. 

Many wayfinding systems use the well-known A* algorithm or its variants for path planning \cite{nguyen2017developing,yao2010path,duchovn2014path}. For example, the work presented by Apostolopoulos \textit{et al.} \cite{apostolopoulos2014integrated} proposes an indoor localization and navigation system for the visually impaired using the embedded smartphone sensors, such as accelerometers and compasses. Their method uses the feedback from human users on the availability of landmarks along the provided path. The system provides the users with audio instructions to reach the desired destination. They utilize the A* algorithm to compute the shortest path. The computational time of approaches that utilize A* is high. Moreover, A* based approaches are not good to handle narrow way-paths which are common in indoor environments. They also heavily rely on the availability of full prior knowledge of the locations and environment.

Chen \textit{et al.} \cite{chen2017mobility} propose a path planning system for emergency guidance in indoor environments based on IoT technologies. They use the statistical properties of mobility of groups of people to provide an individualized path for each group. In this work, a graph of the corridors, doors, and exits is optimized to minimize the total evacuation time for all groups. They implemented their work by utilizing the iBeacon technology and smartphones.

GANs have been used widely for visual data. The most related work to path planning is reported by Hirose \textit{et al.}~\cite{hirose2017go} in which GANs are used to classify images as safe to go to or not for robot navigation purposes. In that work, authors used the observed scene by a robot's camera and classified it as ``GO" or ``NO GO." 

Compared to the aforementioned works, our proposed approach uses a generative deep learning technique over the crowd-sourced trajectory data to suggest a viable path between a given source-destination pair where no pre-recorded path between the given source-destination pair is found in the training dataset. While most of the reported works on GANs rely on vision tasks, our approach is based on non-visual data for path planning which has a high potential for IoT applications as detailed later in Section \ref{sec:conclusion}. 

\section{Proposed Method}\label{sec:method}

Our proposed path planning approach is based on GANs. GANs have been shown to perform impressively in computer vision applications. Especially, it can generate dynamic pictures from static ones, predicting several seconds of a movie, arithmetic on face \cite{radford2015unsupervised}, and text to image synthesis \cite{reed2016generative}. These accomplishments motivate us to consider the use of GANs to generate reliable and correct paths in support of smart mobility applications. 

\subsection{Background on GAN}

Generative adversarial networks \cite{goodfellow2014generative} is a new model of DNNs that has been introduced in 2014. The original model consists of two neural networks that compete against each other. One network is a generator and the other is a discriminator (Cf. Fig. \ref{fig:gan}). The generator network tries to generate samples that resemble the real data such that the discriminator cannot tell whether it is a real sample or a fake one. 

In a GAN framework, a generator model $G$ and a discriminator model $D$ are trained simultaneously. The generative model infers the data distribution $p_g$ over data space $x$. It also includes a defined prior input noise variables $p_z(z)$. The generator is then represented by $G(z; \theta_g)$ that maps $z$ to the data space. $G$ is a differentiable function represented by a DNN with parameters $\theta_g$. The discriminative model $D$ computes the probability that a given sample is coming from the training data or just generated by the generator. The discriminator $D$ is represented by another DNN as $D(x; \theta_d)$ in which its single scalar output specifies the probability that $x$ is available in the real data or is generated using $p_g$.

In the training process of GAN, $D$ is optimized to assign the correct label of real/fake to both training examples and samples from $G$. At the same time, $G$ is optimized to minimize $\log{(1 - D(G(z)))}$. To formally state it, $D$ and $G$ play the following two-player minimax game with value function $V (D, G)$:

\begin{equation}
\begin{gathered}
\min_{G}\max_{D}{V (D, G)} = \mathbb{E}_{x \sim p_{data(x)}} [\log{D(x)}] +\\ \mathbb{E}_{z \sim p_z(z)}[\log{(1 - D(G(z)))}].
\end{gathered}
\end{equation}

The weights of discriminator network are updated by:
\begin{equation}\label{eq:D_update}
\Delta_{\theta_D}\frac{1}{m}\sum_{i=1}^{m}{[\log{D(x_i)} + log{(1-D(G(z_i)))}]}
\end{equation}

while the generator's weights are updated by 
\begin{equation}\label{eq:G_update}
\Delta_{\theta_G}\frac{1}{m}\sum_{i=1}^{m}{\log{(1-D(G(z_i)))}}
\end{equation}

\subsection{Using GANs for Path Planning}

In our work, the map of the environment is specified by a matrix $M = [a_{ij}]; 0 \leq i \leq I; 0 \leq j \leq J $, where $I$ and $J$ are the maximum indexes of vertical and horizontal positions, respectively. $a_{ij}$ equals $1$ represents that $(i,j)$ is in a public access area. Otherwise, a $0$ indicates that it is not a public access area; hence, no path should go through it.
Each path in the training data is also represented by a matrix $P$ as $P = [t_{ij}]$ that has the same dimensions as $M$. In this matrix, $t_{ij}$ equals $1$ indicates that position $(i,j)$ is included in the path. Otherwise, $0$ indicates that it is not included in the path.

\begin{figure}
	\begin{center}		
		\includegraphics[width=.45\textwidth]{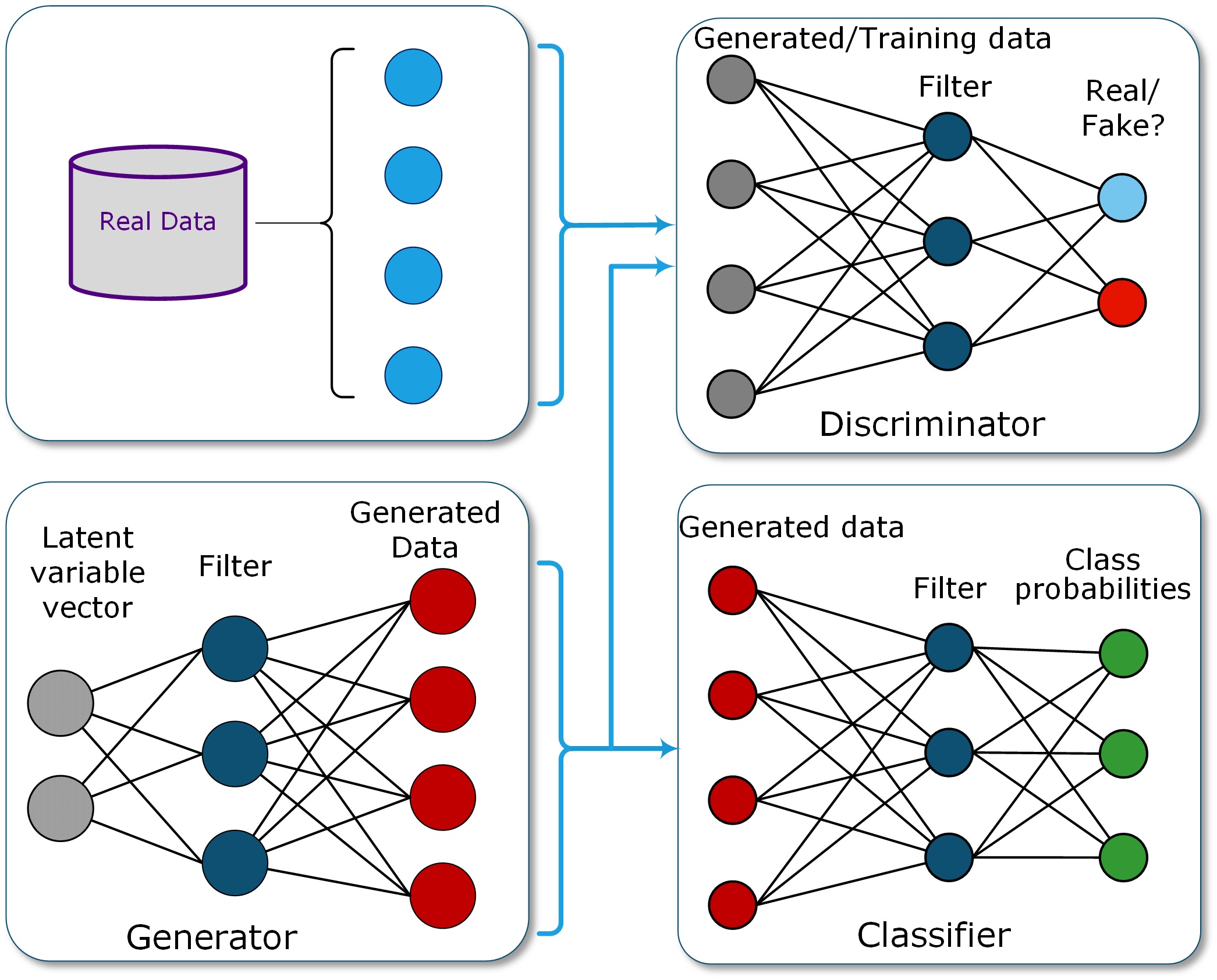}

	\end{center}	
	\caption{Our GAN Based Path Planning Approach.}\label{fig:gan}	
\end{figure}

\begin{figure}
	\begin{center}		
		\includegraphics[width=.45\textwidth]{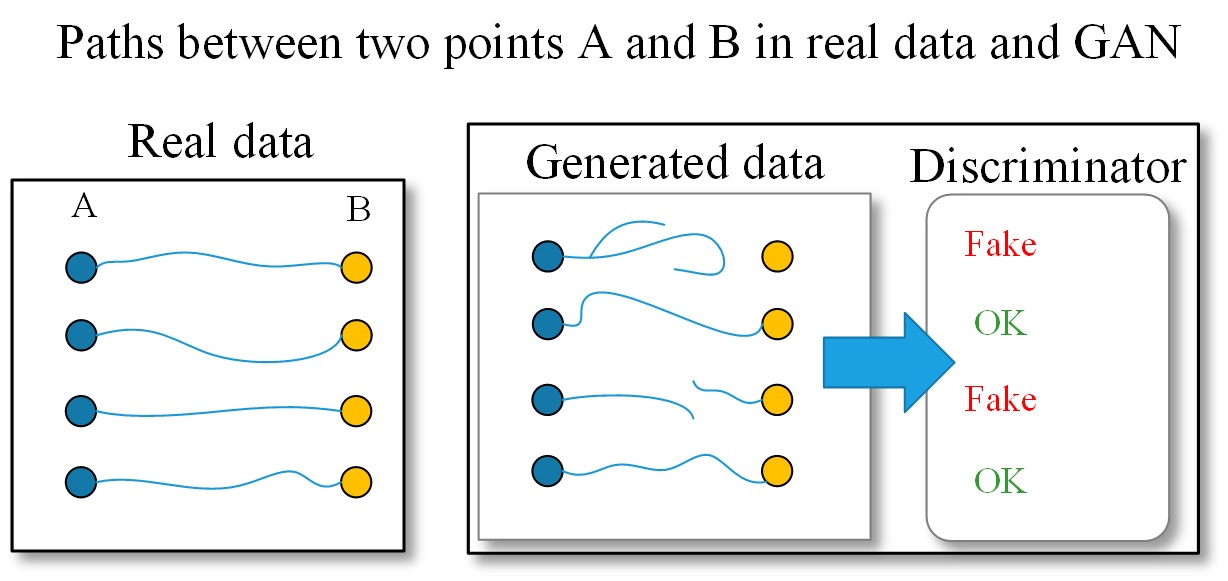}

	\end{center}	
	\caption{The conceptual diagram of using GAN in wayfinding.}\label{fig:gan_use}	
\end{figure}

In path planning services, GANs can be used to generate a path based on the previously collected user trajectories (i.e., real-world data) that lead the users to earlier destinations. Intuitively, the generator network tries to generate a path between a given source-destination pair that seems more realistic rather than a generated fake path. On the other hand, the discriminator network infers the probability of that path being real or not – e.g., a real path should be gradually leading to the destination.  After training the GAN, during the inference phase, the output of the generator network is fed to a classifier network (Cf. Fig. \ref{fig:gan}) to identify the label of the path (i.e., the source and destination of the generated path). Figure~\ref{fig:gan_use} provides a conceptual illustration of using GAN in support of wayfinding.

In our approach, the generator $G$ is a series of fully-connected layers that transforms a noise vector $z$ into a potential path (i.e., in the form of a layout frame that specifies the path). Likewise, the discriminator $D$ is a series of fully connected layers that accepts a layout frame and specifies its probability of being a fake or a real sample. The implementation of our model is modified based on an implementation of GAN~\cite{eriklindernoren2017KerasGAN} using Keras.

Algorithm \ref{alg:gan_path_alg} illustrates the overall process for finding a path by generating a viable path and then classifying it. The algorithm first trains the generator and discriminator of the GAN model simultaneously. A classifier is also trained on the same dataset to classify the generated paths. The GAN, then generates a path until the classifier determines that the generated path complies with the user's request. The alternative approaches for classifying GAN's generated data are presented in \cite{mirza2014conditional, salimans2016improved, odena2016semi} where a discriminator network is extended to classify its inputs that come from real or generated data.  

\begin{algorithm}

\caption{Path Planning Algorithm using GAN}
\label{alg:gan_path_alg}
\begin{algorithmic}[]
\STATE{\textbf{Input:} (\textit{Source},\textit{Destination}), Training dataset}
\STATE {\texttt{/* Training the GAN in \textit{I} iterations*/}}
\FOR {$k = 1$ to $I$ }
	\STATE {Draw $m$ noise samples from $p_g(z)$}
    \STATE {Draw $m$ real examples from $p_d(x)$}
    \STATE {Update the discriminator $D$ based on (\ref{eq:D_update})}
    \STATE {Draw $m$ noise samples from $p_g(z)$}
    \STATE {Update the generator $G$ based on (\ref{eq:G_update})}
\ENDFOR
\STATE {\textbf{end for}}
\STATE {\texttt{/* Finding the path */}}
\STATE {Train a classifier \textit{C} on training data}
\STATE $y \leftarrow (Source, Destination)$
\STATE $\hat{y} \leftarrow None$
\WHILE {$\hat{y} == None$}
\STATE {Generate a path $h$ using the trained GAN}
\STATE {$\hat{y} \leftarrow C(h)$}
\IF {$\hat{y} == y$}
\STATE {return $h$}
\ELSE \STATE{$\hat{y} \leftarrow None$}
\ENDIF
\ENDWHILE 
\end{algorithmic}
\end{algorithm}

\section{Use Case: Wayfinding for Disabled and Visually Impaired People}\label{sec:usecase}

There are more than four million Americans with vision impairments \cite{legge2016indoor}. Assisting the Blind and Visually Impaired (BVI) to navigate through indoor and outdoor environments can contribute to having this important segment of the society live more independently. To provide reliable and accurate guidance to BVI commuters, a precise wayfinding mechanism is needed. There are many attempts in the recent literature to tackle this problem using a variety of technologies. However, these attempts impose new challenges for the users as their accuracy may not be fine-grained. Existing research studies that investigate wayfinding approaches for the BVI (regardless of the underlying technology, such as RFID, sensor networks, or vision) utilize training datasets for the system \cite{ledlie2012mole}. For large deployments, providing such training dataset that needs to be labeled (i.e., class labels) is not always feasible. Instead, exploiting the availability of crowd-sourcing data that users create and manually label, is a solution that expedites the development of such services. In this research, we utilize the experiences of other users and their trajectories to improve the accuracy of the path planning system. 

\subsection{Experimental Setup}\label{ssec:implementation}

We deploy a grid of $13$ iBeacons \cite{newman2014apple} to experiment with our proposed path planning and wayfinding approach in a campus library setting. Our experiments cover the first floor of Waldo Library, Western Michigan University (WMU). In our work, we use the iBeacons' Received Signal Strength Indicator (RSSI) values to serve as the raw source of input data to identify users' locations. iBeacons are installed on the public access areas, so that we experiment with our proposed path planning and wayfinding approach in the coverage area of the iBeacons. Smartphones are also utilized to sense the iBeacons' signals and to compute the current position of the user relative to the known iBeacon positions.

\begin{table*}[]
\centering
\caption{Comparing the coverage range of different communication technologies}
\label{tbl:commTech}
\begin{tabular}{lcccc}
\hline
\multicolumn{1}{c}{\textbf{Technology}} & \textbf{Range} & \textbf{Topology} & \textbf{Pros} & \textbf{Cons}  \\ \hline
RFID                                    & 10 cm $\sim$ 2 m & Peer-to-peer & Easy installment & A dense deployment causes inference \\ \hline
NFC                                     & $\sim$ 10 cm   & Peer-to-peer & Cheap tags & Not widely available in mobile phones   \\ \hline
BLE                                     & $\sim$ 30 m    & Star & Low power; Easy set up & Need fixed reference points  \\ \hline
ZigBee                                  & 10 $\sim$ 20 m & Star/mesh & Low power consumption & Requires pairing devices \\ \hline
Z-Wave                                  & $\sim$ 30 m  & Mesh & High level of interoperability & More expensive devices  \\ \hline
UWB                                     & $\sim$ 100 m & Star & Significant obstacles penetration capacity & Interference with nearby UWB systems  \\ \hline
WiFi                                    & $\sim$ 100 m  & Star/mesh & No need for extra equipment & Not available anywhere  \\ \hline
\end{tabular}
\end{table*}

iBeacons transmit short-range BLE radio waves that consume tiny power. iBeacons can operate for several months or even years. Each iBeacon covers up to ten times the distance of the classic Bluetooth while its transmission is 15 times faster. iBeacons can be configured to transmit at a power level that ranges from 0.01 mW to 10 mW. Moreover, the BLE standard has been developed and adopted by many smartphone makers and is now available on most smartphone models. These characteristics make BLE a good and practical choice for IoT applications that require short-range communications. 
Table \ref{tbl:commTech} summarizes and compares several communications technologies and their advantages and disadvantages. 


In our implementation and experiments with an indoor location-aware application, we chose the iBeacon technology due to its low price and ease of deployment. These iBeacons can be stuck to a wall or ceiling without tapping into the buildings’ infrastructure. iBeacon is a standard technology based on Bluetooth low energy or Bluetooth Smart that was proposed by Apple Inc. iBeacon devices are identifiable and addressable through three separate fields; namely, UUID, major, and minor. 

\begin{figure}
	\begin{center}		
		\includegraphics[width=.45\textwidth]{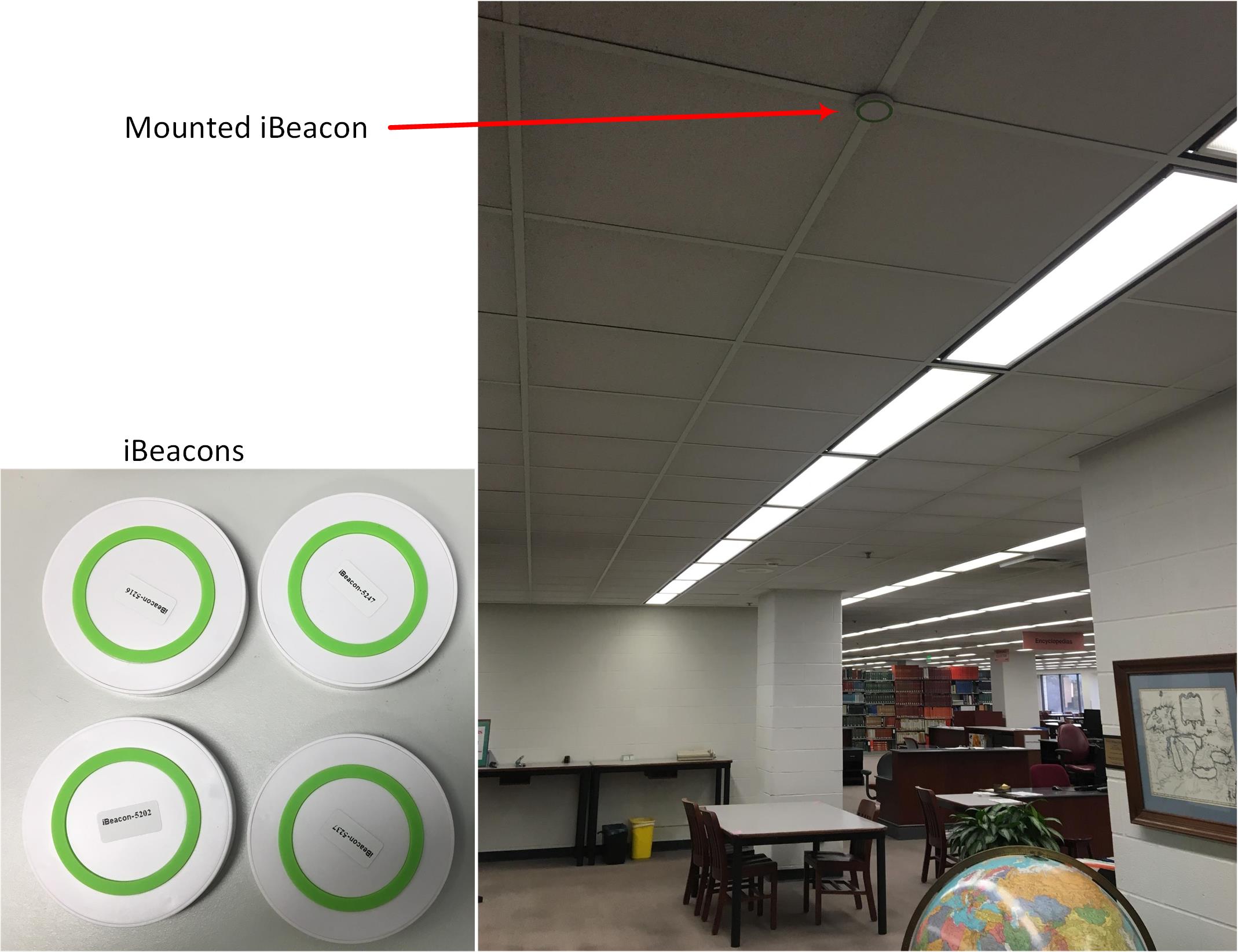}

	\end{center}	
	\caption{iBeacon setup in the environment.}\label{fig:beacon_setup}	
\end{figure}

Figure \ref{fig:beacon_setup} shows a scene of a mounted iBeacon in our area of experiments. iBeacons cannot trigger a location-based action on their own since they are half-duplex transmitting devices that only send their probe signals in a specific interval (usually every one second). So, there is a need for a receiving device that can detect the iBeacons’ signals. Smartphones nowadays are equipped with the BLE technology and are ideal for act as the receivers since they are pervasive. 

\subsection{Our GAN Configuration}\label{ssec:our_gan_config}

We implemented our GAN based approach with the following structure. The generator is a feed-forward DNN with the input layer composed of $100$ neurons, and four hidden layers composed of $256$, $512$, $1024$, $247$ neurons, respectively with a \textit{tanh} activation function. The last layer is composed of $247$ (i.e., $19*13$) neurons which is the dimensionality of the path frames. The discriminator is also a feed-forward DNN with an input layer composed of $247$ neurons and three hidden layers composed of $512$, $256$, $1$ neurons, respectively with a \textit{sigmoid} activation function. The loss function for the model is \textit{binary crossentropy} and the optimization method is \textit{Adam}. The path classifier is also a neural network with an input layer composed of $247$ neurons and two hidden layers that are composed of $256$ and $6$ neurons, respectively. We used the \textit{softmax} as the activation function for that last layer to produce the probability of the $6$ classes. 

\section{Evaluation and Results}\label{sec:evaluation}

In this section, we discuss the datasets used to conduct this research and the evaluation of the proposed system according to the use case in section \ref{sec:usecase}. The evaluations have been done on a computer with $2.3$ GHz Intel Core i5 CPU and $8$ GB RAM.

\begin{figure}
	\begin{center}		
		\includegraphics[width=.5\textwidth]{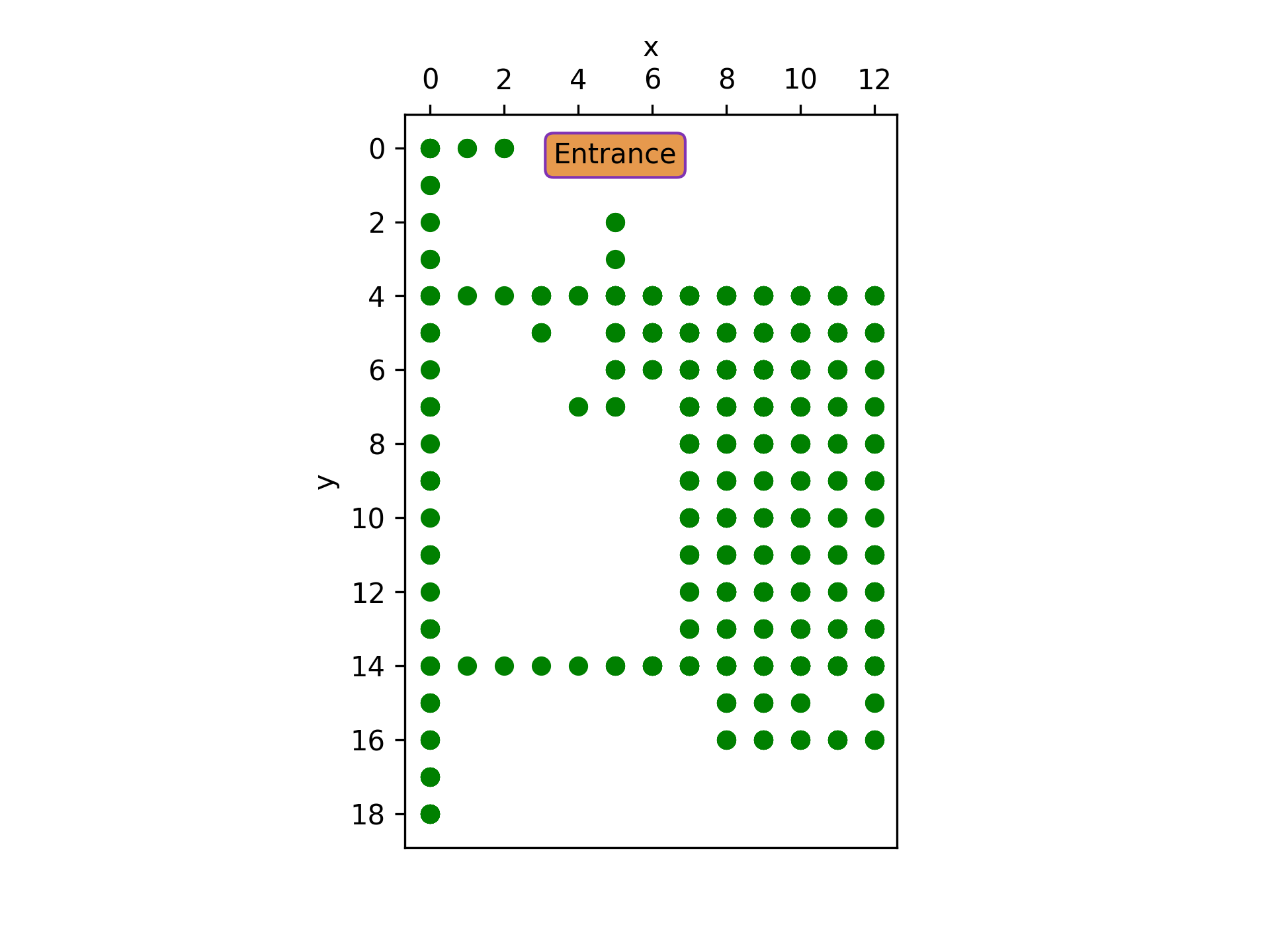}

	\end{center}	
	\caption{Accessible areas for positioning and navigation. The empty spaces represent rooms and other facilities that are not used for public.}\label{fig:allpos}	
\end{figure}

\subsection{Datasets}

Since, our research involves two primary components; namely, localization, and wayfinding, we performed evaluations focusing on the accuracy of these two components on two datasets (i.e.,  localization dataset \cite{blerssi2018mohammadi} and a path planning dataset \cite{pathplanning2018mohammadi}). We deployed the physical infrastructure needed on the first floor of Waldo Library at Western Michigan University to collect these two datasets. Figure \ref{fig:allpos} shows the active area covered by the deployed iBeacons and is accessible to the public. All our localization and navigation experiments are conducted in this area. Also, the localization and path planning datasets are collected from this experimental testbed. 

\textbf{Localization Dataset.} The localization dataset contains $1420$ labeled samples of which we used $70\%$ for training and the remaining $30\%$ for testing. Each sample consists of RSSI values of the $13$ iBeacons deployed on this testbed. Each of which is presented as a feature. In addition, two positioning features representing the \texttt{x} and \texttt{y} coordinates of the current location serve as predictable features.

\begin{figure}
	\begin{center}		
		\includegraphics[width=.40\textwidth]{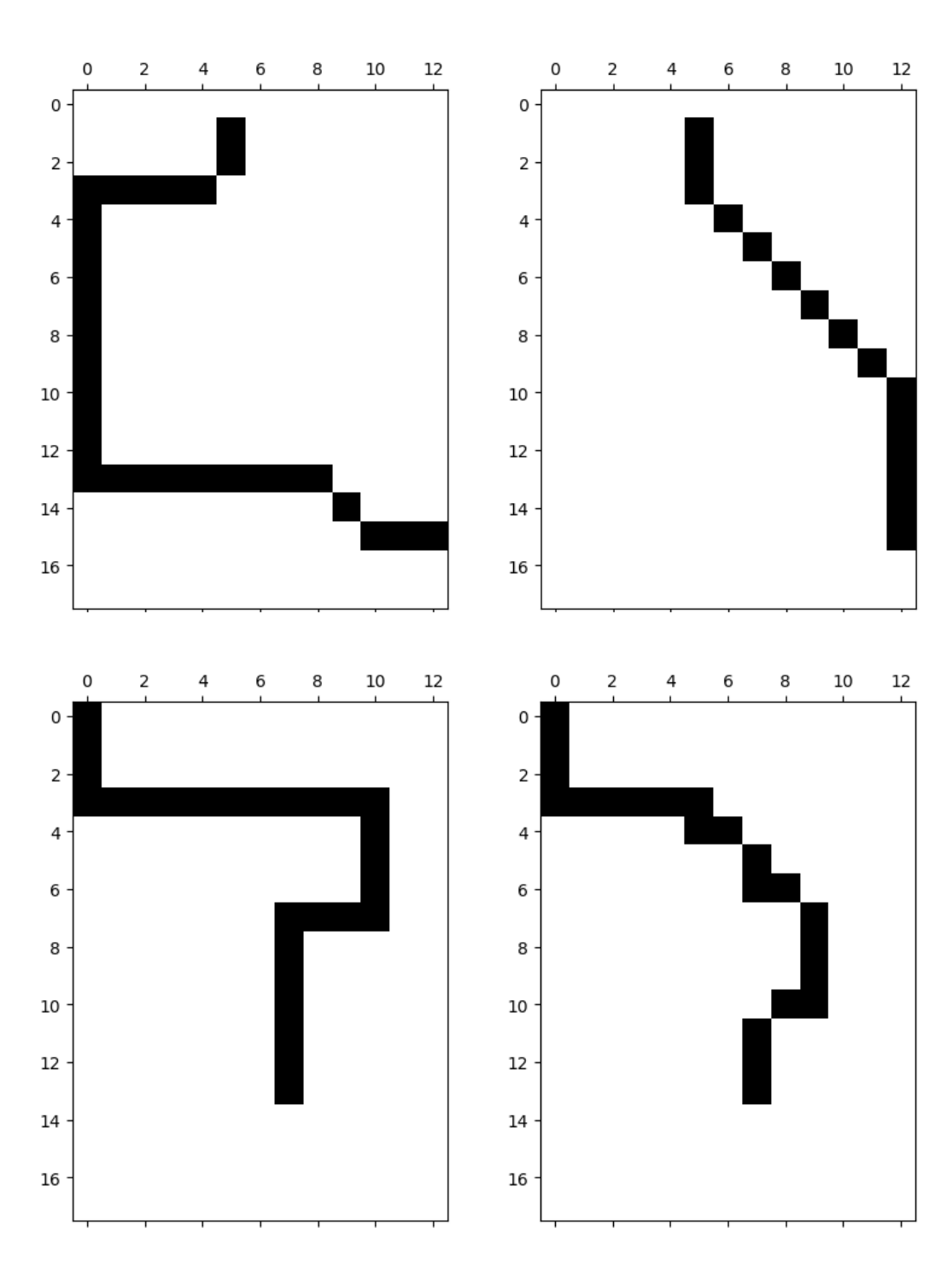}
	\end{center}	
	\caption{Samples of paths in training dataset. The top figures represent class 2 (from (2, 5) to (16, 12)) and the bottom figures represent class 0 (from (0, 0) to (14, 7)).}\label{fig:paths}	
\end{figure}

\textbf{Path Planning Dataset.} In the path planning dataset, we have $6$ different classes that define the source and destination of a path. Table \ref{tbl:path_classes} shows the different classes of the paths and their corresponding source and destination coordinates. Each class has about $52$ samples with a total of $313$ samples that are stored as CSV files. Each file resembles a frame containing the path in a given indoor environment. A path frame is a $19\times13$ matrix in which the item at indexes $(i, j)$ indicates that the corresponding position of $(i, j)$ part of the path (value at indexes $(i, j)$ equals $1$) or not (value at indexes $(i, j)$ equals $0$). Figure \ref{fig:paths} depicts some paths from the training dataset.

\begin{table}
\centering
\caption{Coding the Paths Labels}
\label{tbl:path_classes}
\begin{tabular}{|l|l|l|l|l|}
\hline
\multirow{2}{*}{\textbf{Class}} & \multicolumn{2}{l|}{\textbf{Source}} & \multicolumn{2}{l|}{\textbf{Destination}} \\ \cline{2-5} 
                       & \textbf{x1}           & \textbf{y1}           & \textbf{x2}              & \textbf{y2}             \\ \hline \hline
0                      & 0            & 0            & 14              & 7              \\ \hline
1                      & 0            & 0            & 16              & 12             \\ \hline
2                      & 2            & 5            & 16              & 12             \\ \hline
3                      & 2            & 5            & 16              & 8              \\ \hline
4                      & 2            & 5            & 18              & 0              \\ \hline
5                      & 4            & 0            & 10              & 10             \\ \hline
\end{tabular}
\end{table}

\subsection{Localization}

The training process of localization is a multi-label classification problem since we need to predict \texttt{x} and \texttt{y} coordinates. We used the binary relevance approach \cite{read2011classifier} to perform training and classification of RSSI values into location coordinates. In this approach, for each class feature (two features in our problem; namely, \texttt{x} and \texttt{y} coordinates), a separate model is created and trained on the same training dataset. 

\begin{figure}
	\begin{center}		
		\includegraphics[width=.45\textwidth]{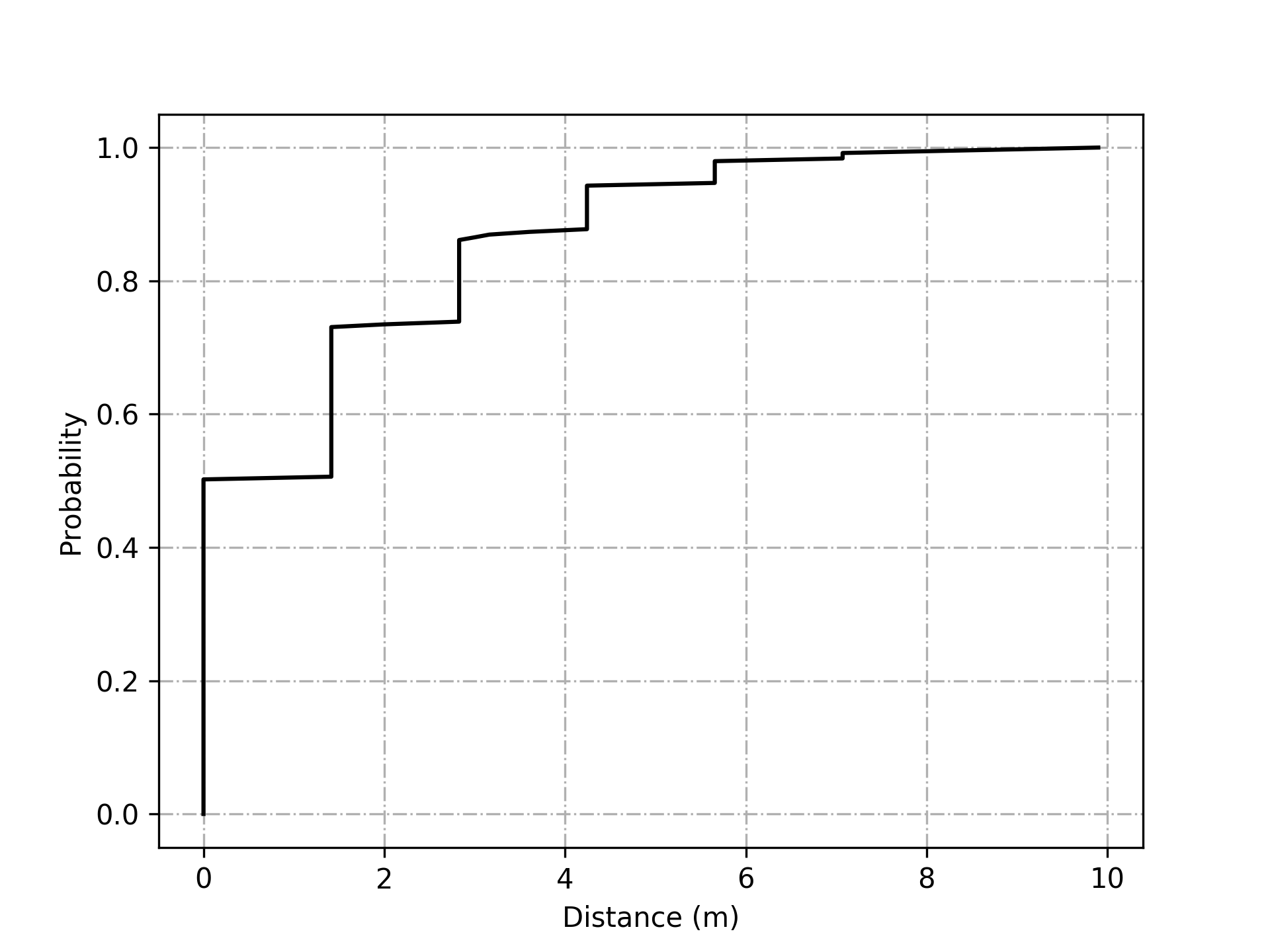}

	\end{center}	
	\caption{Cumulative Density Function (CDF) of Euclidean distance error for localization.}\label{fig:dist_error}	
\end{figure}

Figure \ref{fig:dist_error} illustrates the accuracy of the localization method in terms of the Euclidean distance error. The plot shows that about $50\%$ of the results are an exact match. Also, more than $75\%$ of targets have a distance error of $2$ m or less. The average error is $1.5$ m. The average prediction time per sample for localization is $0.12$ millisecond. 

\subsection{Path Planning}

Assessing the accuracy of a generative model is challenging since there is no standard metric that quantifies to what degree the output is realistic. The metric also needs to specify how novel the output is compared to the real data to avoid having a model that just remembers the input training data and provides it again as output \cite{alzantot2017sensegen}.  The research work on generative models in the context of vision-based tasks where images are generated utilize human judgments to assess the quality of output samples (e.g., mean opinion score \cite{ledig2017photo}).

\begin{figure}
	\begin{center}		
		\includegraphics[width=.40\textwidth]{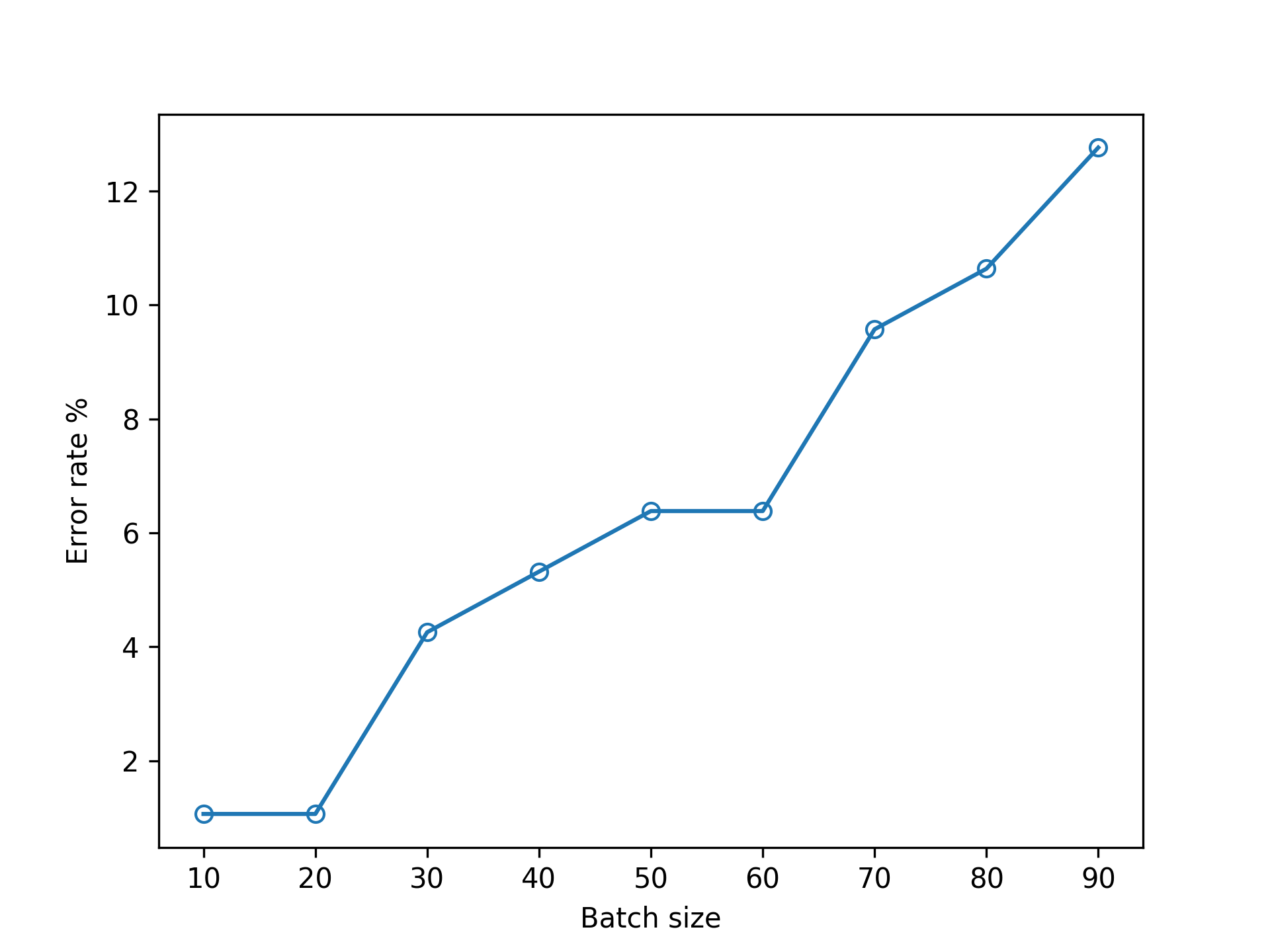}

	\end{center}	
	\caption{Error rate of the path classifier when training is performed with different batch sizes.}\label{fig:pathClass_error}	
\end{figure}

We evaluated the path classifier separately, since we need ground truth values to examine how good it can identify the classes of unseen paths. To evaluate the accuracy of the path classifier, we split the path planning dataset into $70\%$ for training and $30\%$ for the testing. The average error rate of the classifier is $1.06\%$. 
Figure~\ref{fig:pathClass_error} demonstrates the error rate of the path classifier when we set different batch sizes for training. We get the best accuracy when the batch size is $20$ or less. That is because with smaller batch sizes, Stochastic Gradient Descent (SGD) based methods use smaller steps in the training process to find the optimum, while larger batch sizes indicate larger steps. So, the larger steps may skip some local optima and degrade the quality of the model~\cite{keskar2016large}. 

\begin{figure}
	\begin{center}		
		\includegraphics[width=.45\textwidth]{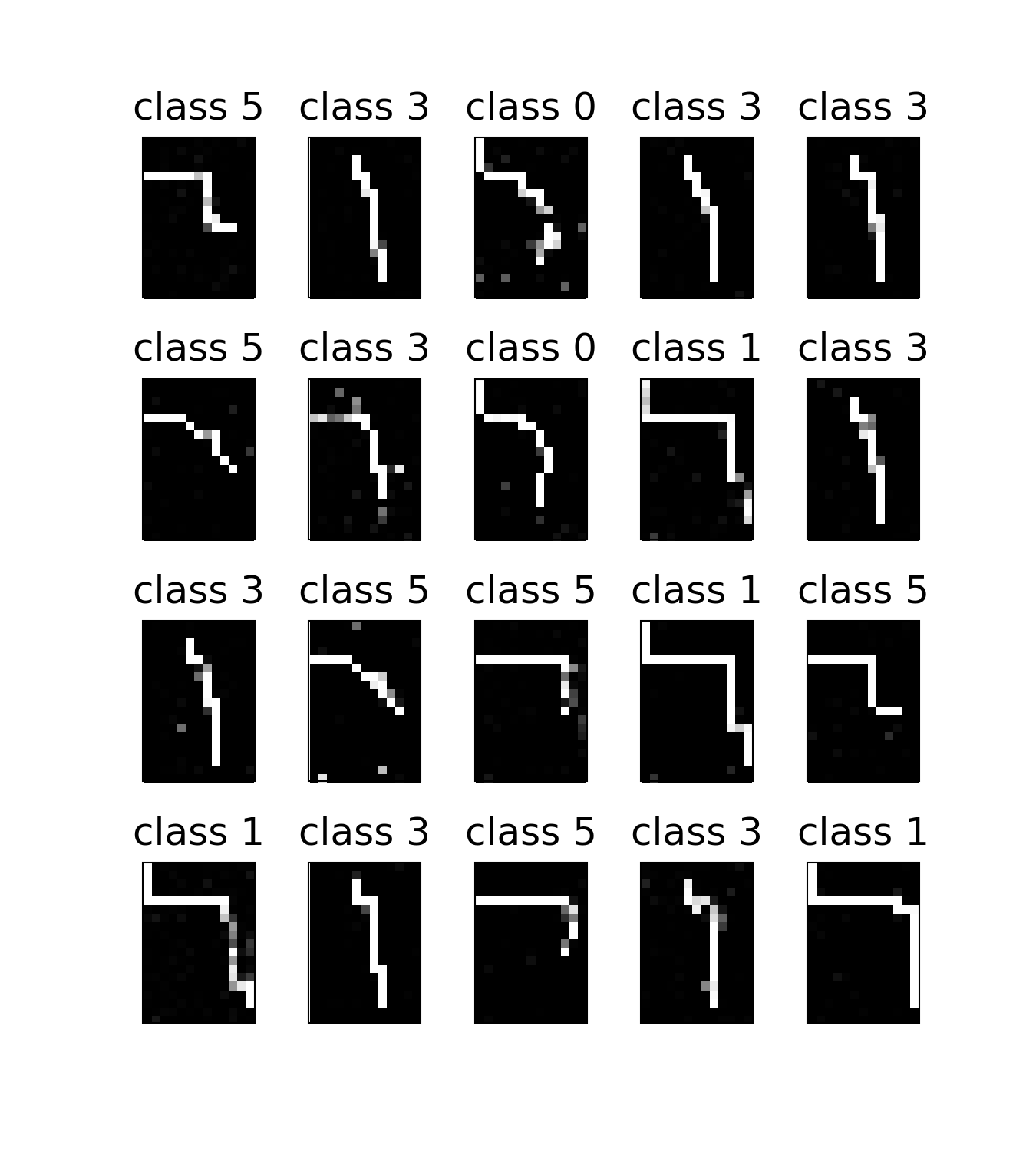}
	\end{center}	
	\caption{Samples of randomly generated paths and their classification.}\label{fig:gen_paths}	
\end{figure}

\begin{figure}
	\begin{center}		
		\includegraphics[width=.40\textwidth]{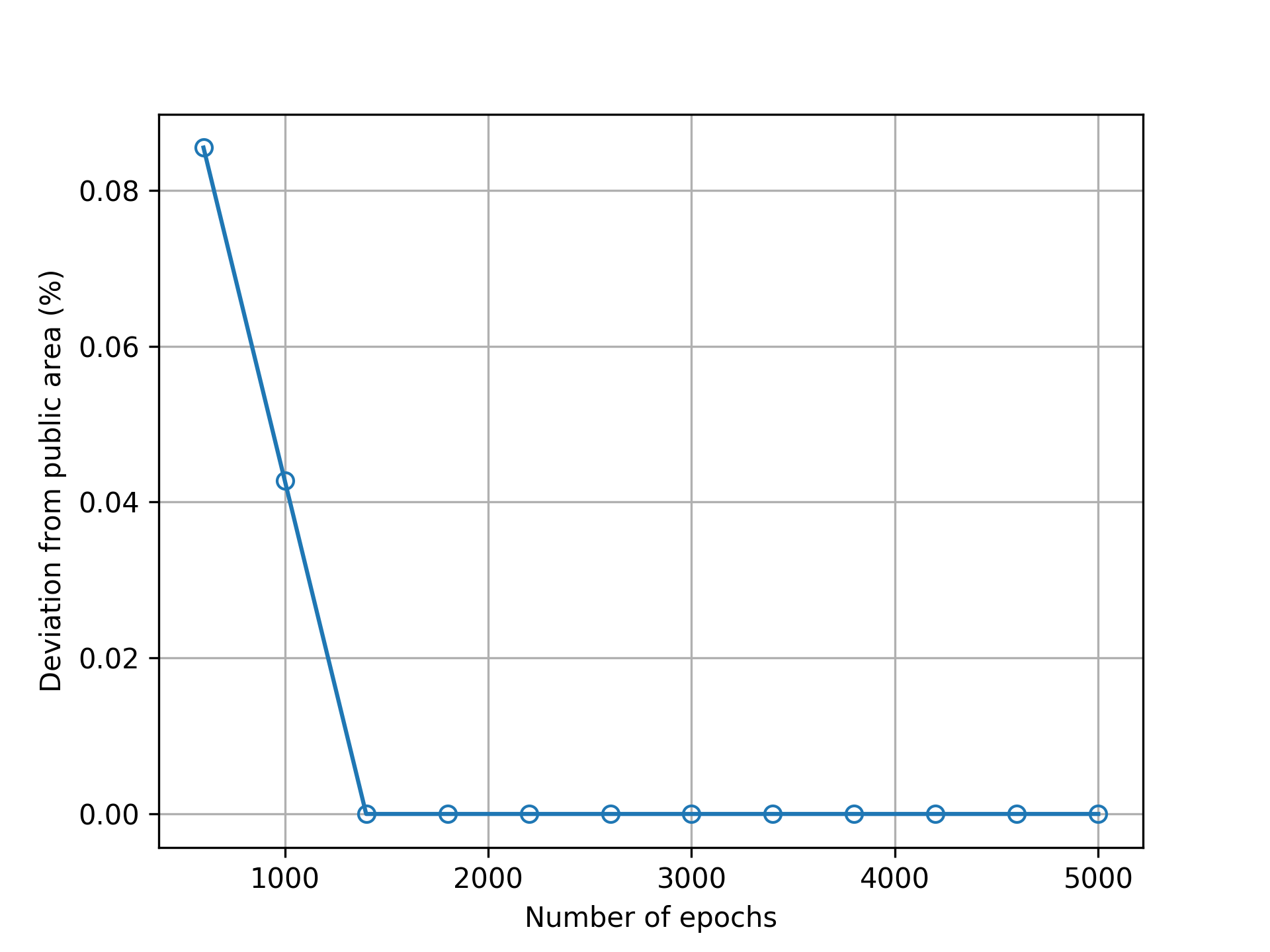}
	\end{center}	
	\caption{Deviation from the public access area for the generated paths when the model is trained using different number of epochs.}\label{fig:deviation}	
\end{figure}

Figure \ref{fig:gen_paths} shows several paths generated using our GAN based approach and their corresponding labels. There are some generated samples with minor noise which can be filtered in a production setting. One interesting evaluation of this work is to see how the generated paths may deviate from the baseline public access area (Cf. figure~\ref{fig:allpos}). We assessed this by considering those positions in the generated paths that are outside of the public access areas. This can be formulated as:
\begin{equation}
Score_{deviation}(G) = \frac{|G| - |G \cap B|}{|B|},
\end{equation}
where $G$ is a set of positions that represent the generated path and $B$ is the set of the positions for the public access area. This measurement is an indication of the ``safety" of the paths. In other words, it points out how much the generated paths are inside the safe boundaries (e.g., do not go to the restricted areas, etc.). Our experiments show that the deviation is minor if any and in the range of $0$ to $0.08\%$. Figure \ref{fig:deviation} shows the deviation percentage of the generated paths based on the model that is trained in a different number of epochs. As we train the model with a larger number of epochs beyond $1000$, the results do not exhibit any deviation.

We also conducted a subjective evaluation using the Mean Opinion Score (MOS) to assess the quality of the generated paths. For this evaluation, $60$ generated paths were rated with a score that ranges from $1$ (bad) to $5$ (very good) by three raters based on their clarity to be followed for navigation purposes. The average MOS for all the scored paths is $4.46$ ($89\%$) with a standard deviation of $0.74$. Table \ref{tbl:mos} summarizes these results. These results demonstrate the potential of using the generated paths for reliable wayfinding and navigation tasks.   

The training time for our GAN based approach is $14$ minutes for $15000$ epochs and $48$ minutes for $50K$ epochs. The run time for generating a path instance by GAN took $0.21$ millisecond and the run time for classifying a generated path instance is $0.14$ millisecond. 

\begin{table}[]
\renewcommand{\arraystretch}{1.2}
\centering
\caption{Mean opinion score for the generated paths.}
\label{tbl:mos}
\begin{tabular}{|l|c|c|}
\hline
\textbf{Rater}\# & \textbf{MOS} & \textbf{standard deviation} \\ \hline \hline
1     &  4.6   &  0.66   \\ \hline
2     &  4.7   &  0.45   \\ \hline
3     &  4.1   &  0.88   \\ \hline \hline
\textbf{Avg}    &  \textbf{4.46}   &  \textbf{0.74}   \\ \hline
\end{tabular}
\end{table}

\section{Conclusion and Future Work}\label{sec:conclusion}
This paper presented an approach for path planning based on GANs and crowd-sourced data. In our approach, a GAN is used to generate paths to a desired destination. We experimented with the proposed approach in a wayfinding application in an IoT-enabled environment. Our evaluations show that the generated paths are more than $99.9\%$ safe and reliable while the path classifier component correctly classifies the paths with an accuracy of up to $99\%$. Also, the generated paths are of high quality attaining around $89\%$ mean opinion score.

GANs can serve as an efficient technique for generating samples that do not exist in the collected dataset. However, it is hard to adjust their parameters optimally since assessing their performance is hard. In particular, the loss function is not necessarily a good indicator of their goodness. So far, the most trusted method to evaluate their output is to use human judgments~\cite{salimans2016improved}.

A potential pathway for the future work is to provide tactile feedback (e.g., vibration codes) to direct the user and offer accurate guidance information to ensure that the user follows the planned path. Providing a mechanism to generated paths that minimize or maximize multiple metrics is another direction for future work. Another application where the proposed approach can be utilized is for path planning for autonomous vehicles. In such an application, a vehicle can utilize the path that is generated by a GAN based on the trajectories of other participating vehicles that have shared their GPS encoded paths for training the GAN. 

\section*{Acknowledgment}
This study was partially supported by the USDOT through the Transportation Research Center for Livable Communities (TRCLC), a Tier 1 University Transportation Center at Western Michigan University. 

\bibliographystyle{IEEEtran}

\bibliography{references}

\end{document}